\documentclass[11pt]{article}

\usepackage{amsmath}
\usepackage{amsthm}
\usepackage{amssymb}
\usepackage{rotating}
\usepackage{array}
\usepackage{enumitem}
\usepackage{mathtools}
\usepackage{setspace}
\usepackage{fullpage}
\usepackage{hyperref}
\usepackage{algorithm}
\usepackage{booktabs}
\usepackage{xspace}
\usepackage{authblk}
\usepackage{todonotes}

\newcommand{\BAn}{Barab\'{a}si-Albert\xspace}
\newcommand{\ERn}{Erd\H{o}s-R\'{e}nyi\xspace}
\newcommand{\WSn}{Watts-Strogatz\xspace}

\title{Human intuition as a defense against attribute inference}

\author[a]{Marcin Waniek}
\author[a,*]{Navya Suri}
\author[a,*]{Abdullah Zameek}
\author[b]{Bedoor AlShebli}
\author[a]{Talal Rahwan}

\affil[a]{Computer Science, New York University Abu Dhabi, Abu Dhabi, UAE}
\affil[b]{Social Science, New York University Abu Dhabi, Abu Dhabi, UAE}
\affil[*]{Equal contributions}

\date{}

\begin{document}

\maketitle

\begin{abstract}
Attribute inference---the process of analyzing publicly available data in order to uncover hidden information---has become a major threat to privacy, given the recent technological leap in machine learning. One way to tackle this threat is to strategically modify one's publicly available data in order to keep one's private information hidden from attribute inference. We evaluate people's ability to perform this task, and compare it against algorithms designed for this purpose. 
We focus on three attributes: the gender of the author of a piece of text, the country in which a set of photos was taken, and the link missing from a social network. For each of these attributes, we find that people's effectiveness is inferior to that of AI, especially when it comes to hiding the attribute in question. Moreover, when people are asked to modify the publicly available information in order to hide these attributes, they are less likely to make high-impact modifications compared to AI. This suggests that people are unable to recognize the aspects of the data that are critical to an inference algorithm.
Taken together, our findings highlight the limitations of relying on human intuition to protect privacy in the age of AI, and emphasize the need for algorithmic support to protect private information from attribute inference.
\end{abstract}


\section{Introduction}
\label{sec:introduction}

In recent years, the algorithms that predict our attributes based on publicly available information have reached staggering levels of effectiveness and sophistication. Easy access to vast amounts of high-resolution data has granted AI algorithms almost clairvoyant-like powers. By analyzing our digital footprint an algorithm can judge our personality traits better than our loved ones~\cite{youyou2015computer}, by processing a photo of our face it can uncover our sexual orientation~\cite{wang2018deep}, and by scrutinizing the dynamics of our keystrokes it can infer our emotional state~\cite{epp2011identifying}. While incredibly impressive as a technological achievement, many consider these advancements in prediction techniques deeply unsettling, since the attributes that can easily be inferred by such techniques include sensitive data that can be used against us. For instance, knowledge about personality traits and emotional states can be used to manipulate one's behavior~\cite{buss1987tactics}, while knowledge about sexual orientation can lead to discrimination in certain parts of the world. Mass applications of AI-driven surveillance technologies by authoritarian regimes can significantly strengthen their control over the population~\cite{polyakova2019exporting}. Many fear that living under the ever-watchful eye of artificial intelligence will lead to a new kind of technological dystopia~\cite{harari201821}.

Many of the privacy solutions in the literature are based on the role of a centralized authority. Notions such as $k$-anonimity~\cite{sweeney2002k}, differential privacy~\cite{dwork2008differential}, and federated learning~\cite{li2020federated} work towards securing sensitive information under the assumption that a set of rules will be provided and obeyed by a central governing force. However, real-life institutions can be prone to error, negligence, or even malice, as evidenced by numerous privacy-related scandals in the recent years~\cite{isaak2018user}. One possible solution to this issue would be to put responsibility for privacy protection into the hands of the general public, letting them strategically shape their publicly available data in order to guard the information they deem sensitive.

In this work, we examine the feasibility of people protecting their own privacy from attribute inference, without any kind of algorithmic help. In particular, we consider two main research questions. First, \textit{how effective are people in inferring private attributes, when compared to algorithms?} While privacy protection remains the primary motivation behind our study, understanding people's ability to infer hidden information could help us understand how they think when attempting to hide such information. Second, \textit{how effective are people in hiding private attributes from inference, when compared to algorithms?} If members of the general public are able to conceal information from AI without any algorithmic support, then it would be enough to simply inform them about the potential risk of their sensitive data being uncovered. If, on the other hand, they are incapable of hiding private attributes on their own, it would underscore the need for developing algorithms that can support them in this endeavor.

In more detail, we consider three different attributes that can be inferred based on publicly available data: the gender of the author of a review, the country in which a set of photos was taken, and a missing link in a social network. For each of these attributes, we compare the performance of participants against AI algorithms in two tasks: inferring the attribute, as well as preventing its inference (i.e., modifying the publicly available data in order to make it harder for algorithms to infer the private attribute). Our results help us understand people's ability to take the safety of their sensitive information into their own hands.

\section{Results}
\label{sec:results}

\begin{figure}[t!]
\centering
\includegraphics[width=\linewidth]{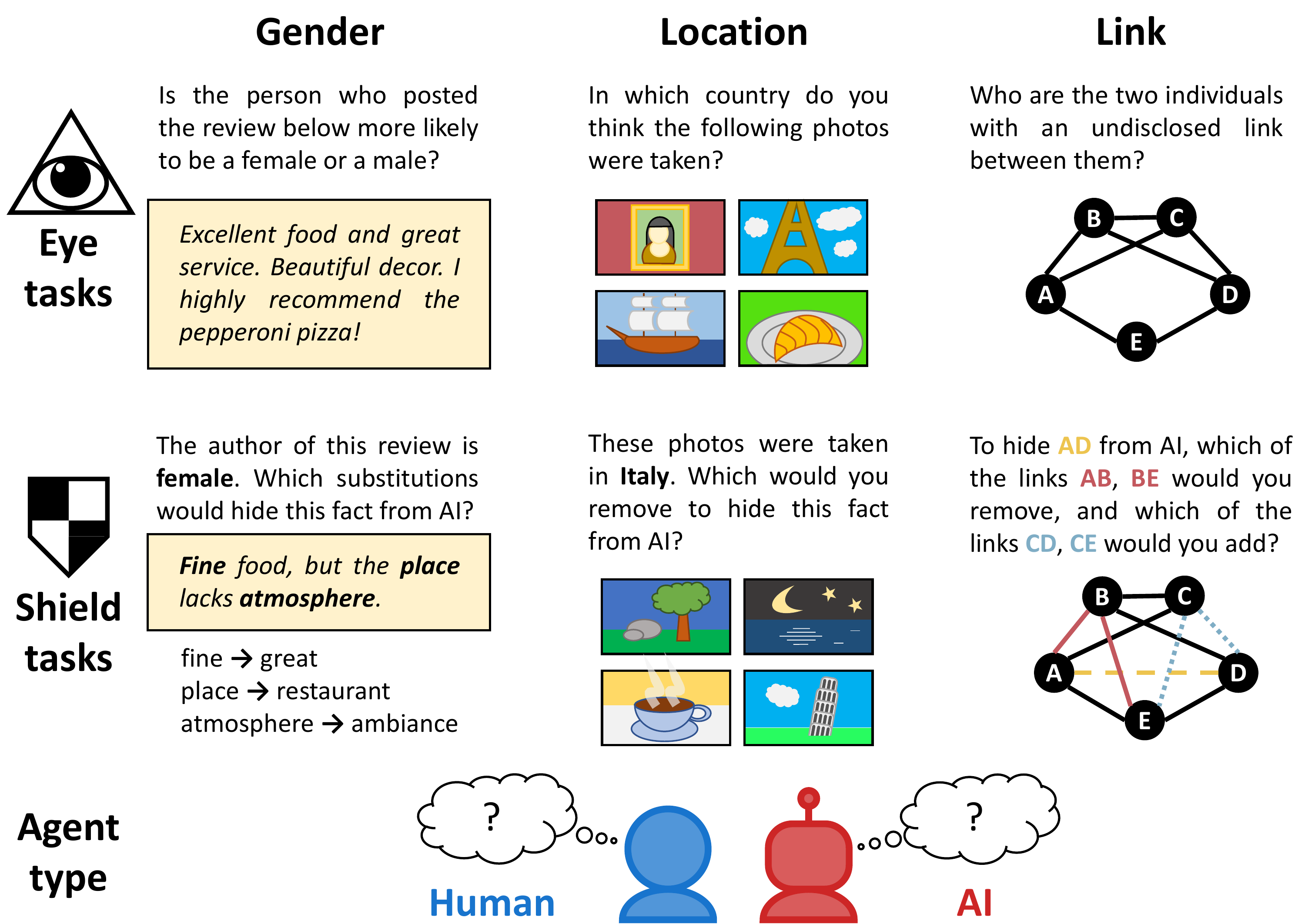}
\caption{
\textbf{The general outline of our experiment.} We focus on three attributes: (i) the \textit{gender} of the author based on the text of the review, (ii) the \textit{location} of origin based on a set of pictures, and (iii) the undisclosed \textit{link} based on the structure of a social network. For each of these attributes, we consider the \textit{eye task}, which involves inferring the attribute based on available data, and the \textit{shield task}, which involves modifying the data in order to make it harder for an AI algorithm to infer the attribute. Each of the six tasks is given to two types of agents: people (participants recruited online), and AI (algorithms trained on data), in order to compare their performance.
}
\label{fig:infographic}
\end{figure}

Our analysis focuses on three attributes: (i) the \textit{gender} of the author of a particular piece of text, (ii) the \textit{location} in which a particular set of photos was taken, and (iii) the undisclosed \textit{link} in a particular social network. For each of these attributes, we focus on two distinct tasks, and refer to the entity solving these tasks as an \textit{agent}, which could either be a human or an algorithm. In the first task, the attribute of interest is hidden, and the agent is required to infer this attribute from the given data. For example, in the case of gender, the agent is presented with a piece of text, and is required to infer the author's gender. In the second task, the attribute of interest is revealed, and the agent is required to modify the given data in order to make it harder for an algorithm to infer that attribute. For example, in the case of gender, the agent is presented with a piece of text along with the author's gender, and is asked to modify the text in order to keep author's gender hidden from prediction algorithms. The first task will be referred to as the \textit{eye task}, since involves ``seeing'' hidden information, while the second task will be referred to as the \textit{shield task}, since it involves ``protecting'' hidden information. The general outline of all eye and shield tasks is illustrated in Figure~\ref{fig:infographic}.

We now briefly comment on generating the instances of the tasks. The technical details of the process are presented in the Methods section. For the gender attribute, we generate the corresponding task instances using a dataset of Yelp reviews~\cite{reddy2016obfuscating}. Each eye instance consists of the text of a review. To construct a shield instance, we select four words of the review that are most indicative of the author's gender, as well as four words that are least indicative. We measure how indicative a given word is according to the normalized pointwise mutual information~\cite{church1990word}, the criterion employed by Reddy and Knight~\cite{reddy2016obfuscating}. The agent is then asked to select three of these eight words to be substituted by their synonyms.
For the location attribute, we use a dataset of Flickr photos~\cite{yang2020protecting}. Each eye instance consists of a set of $16$ randomly chosen photos taken in the same country. To construct a shield instance, we identify four photos whose removal yields the largest drop in the location prediction accuracy, as well as four photos whose removal yields the smallest drop. The agent is then asked to select three of these eight photos to be removed.
For the link attribute, we generate networks using three different models: \BAn~\cite{barabasi1999emergence}, \ERn~\cite{erdds1959random}, and \WSn~\cite{watts1998collective}. Each eye instance consists of a network from which we randomly remove one link (this is the undisclosed link that the agent is asked to identify). To construct a shield instance, we identify two links whose removal causes the greatest decrease in the effectiveness of link prediction, as well as two links whose removal causes the least decrease. Moreover, we identify two links whose addition causes the greatest decrease, as well as two whose addition causes the least decrease. The agent is then asked to select three of these eight modifications to be introduced.

We recruited participants using Amazon Mechanical Turk~\cite{mturk}. The online questionnaire is presented in Section~\ref{app:survey-questionnaires} of the Supplementary Materials. Altogether, $1168$ participants solved the comprehension test and completed their tasks. The exact distribution of participants per task is presented in Table~\ref{tab:survey-stats} in the Supplementary Materials. The participants' number was determined using a power analysis of the pilot study. We preregistered our main findings using the AsPredicted.org portal~\cite{aspredicted}.

\begin{figure}[t!]
\centering
\includegraphics[width=\linewidth]{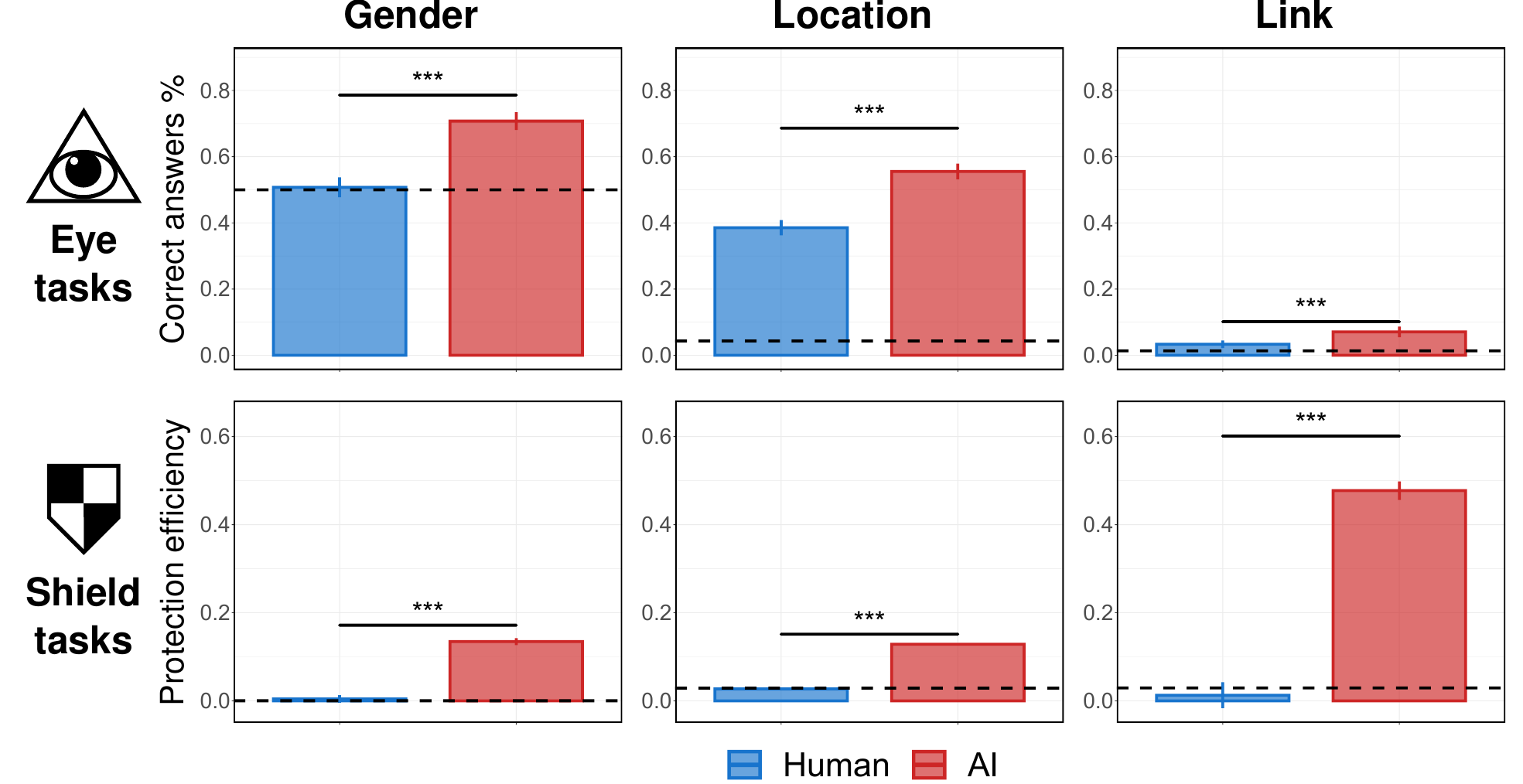}
\caption{
\textbf{Average performance of people vs.~algorithms in the eye tasks and the shield tasks.}
Each column corresponds to a different attribute (gender, location, and link). The first row presents results of the eye tasks with $y$-axes corresponding to the percentage of the correct answers. The second row presents results of the shield tasks, with $y$-axes corresponding to the protection efficiency (see Methods). Each plot compares the average performance of survey participants and AI algorithms in a given task, with the dashed line highlighting the performance of a random baseline. Error bars represent $95\%$ confidence intervals. All results are significant with \textit{p}-values smaller than $0.001$ according to the Welch's \textit{t}-test (the exact values are presented in Table~\ref{tab:survey-stats} in the Supplementary Materials).
}
\label{fig:basic}
\end{figure}

\begin{figure}[t!]
\centering
\includegraphics[width=\linewidth]{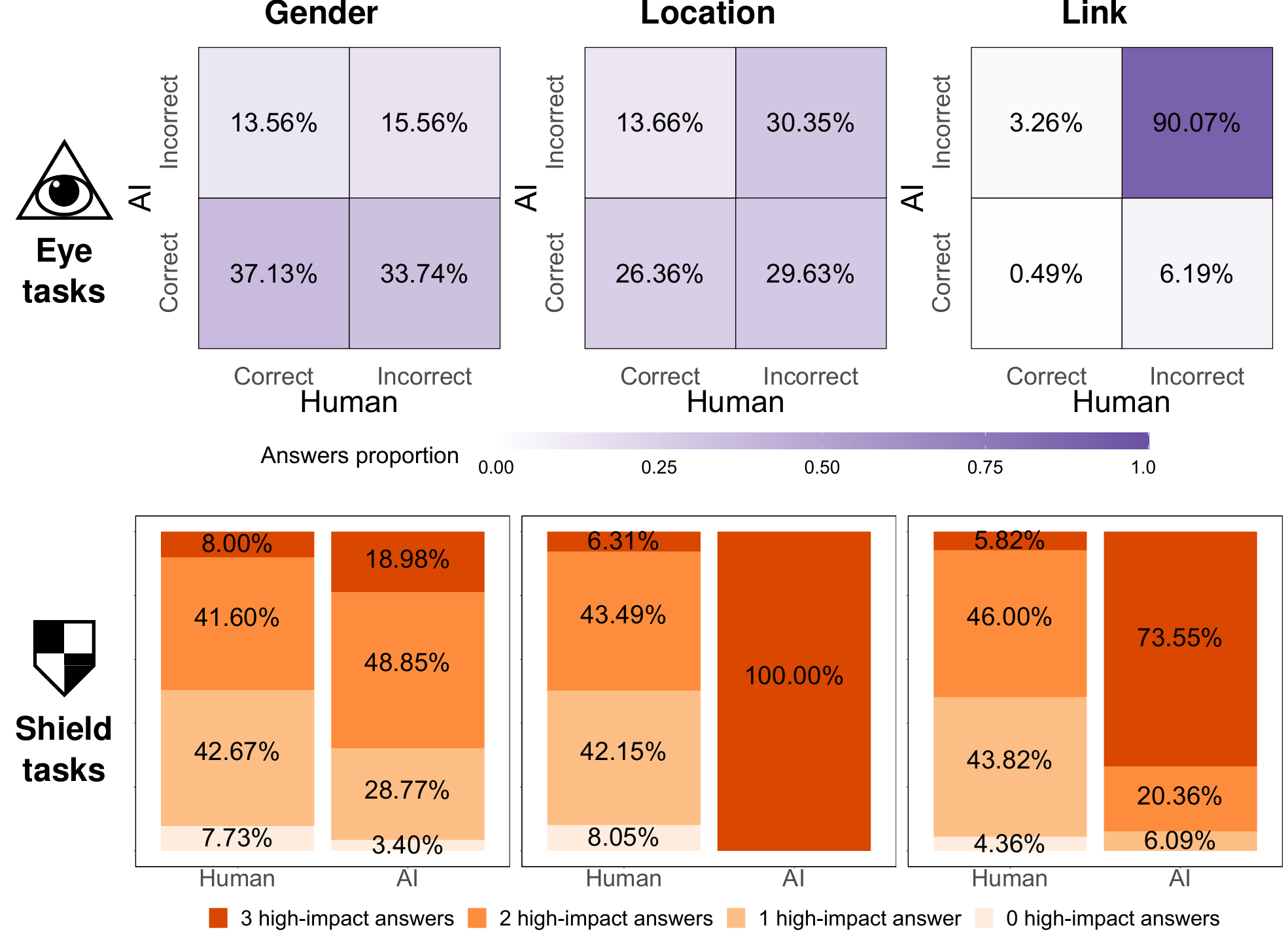}
\caption{
\textbf{Comparison of correct/incorrect and high-impact/low-impact answers in the eye tasks and the shield tasks.}
For any given eye instance, the algorithm assigns probabilities to different answers, and is evaluated based on the one with the highest likelihood of being correct. To provide a similar evaluation to humans, they are evaluated based on the most common answer. The first row focuses on the eye task, comparing the percentage of instances that were correctly and incorrectly answered by the participants and by AI. The second row focuses on the shield task, comparing the percentage of high-impact and low-impact answers selected by the participants and by the AI.
}
\label{fig:correct-high}
\end{figure}

Figure~\ref{fig:basic} compares the performance of participants and AI algorithms in the eye tasks and the shield tasks. As can be seen, AI outperforms humans in every task.
When we focus on the eye tasks, the average performance of both types of agents is the greatest in gender prediction, followed by location prediction, with the average performance in link prediction being the poorest. This is consistent with the number of possible answers to each eye instance, as the agent has to choose one of two genders, one of twenty three countries, or one of about seventy five possible links, respectively.
When we turn our attention to the shields tasks, the average performance of AI is inversely proportional to the performance in the corresponding eye task. In other words, if the problem is already difficult even without any strategic obfuscation, it makes it easier to add an additional layer of confusion. However, the situation is much more dire for the human agents attempting to perform the shield task, as their average performance seems to be comparable to the random baseline.
Altogether, our results present a rather dire perspective of human ability compared to AI. Not only are humans consistently outperformed by algorithms throughout all settings, but their ability is particularly lacking when it comes to privacy protection (i.e., the shield task). These findings underscore the need for both systemic solutions that guard our sensitive information, as well as tools and techniques that would assist us in taking the responsibility for protecting our privacy in our own hands.

Figure~\ref{fig:correct-high} gives us a deeper insight into the differences in performance between humans and AI.
For about half the instances of the gender- and location-based eye tasks, the answers of human and AI agents are either both correct, or both incorrect. The gap in performance results from the distribution of the other half. More specifically, the number of instances in which AI, but not humans, give a correct answer is two to three times greater than those in which humans, but not AI, give a correct answer ($33.74\%$ vs.\ $13.56\%$ for gender, and $29.63\%$ vs.\ $13.66\%$ for location). This difference is the source of the  AI's competitive advantage. As for the link attribute, we observe a much greater percentage of instance where both types of agents give incorrect answers ($90.07\%$). Again, when it comes to the instances in which the outcomes differ, we see about twofold difference in AI's favor ($6.19\%$ vs.\ $3.26\%$). 
We now turn our attention to the shield tasks. Notice that for each of the three attributes, every instance has four of the possible modifications selected as having the greatest impact on prediction quality (we refer to these as the \textit{high-impact} answers), and the other four selected as having the smallest impact (we refer to these as the \textit{low-impact} answers).
The figure shows that when trying to hide private information, people consistently select a smaller percentage of high-impact answers than AI. The difference is especially pronounced in the case of location, as the inference algorithm simply takes a linear combination of the scores of all photos. Thus, an AI algorithm trying to obscure the location information has a relatively easy task of minimizing the accuracy. However, even for the other two attributes, where the prediction algorithms are non-linear, the ability of AI agents to identify high-impact answers is much greater than human agents. This finding suggests that the inability of people to discern what aspects of the given instance are meaningful to the AI may be responsible for their poor performance---a conclusion that is consistent with previous findings from the literature~\cite{waniek2022hiding}.

\section{Discussion}
\label{sec:discussion}

In this work, we compared the ability of human and AI agents to infer private information based on publicly available data, as well as their ability to protect such information against inference attacks. We found that the performance of people is significantly inferior to that of algorithms in both tasks, across all considered attributes. This deficiency of human capability is particularly worrying when protecting one's privacy, as the performance of our study participants is close to the random baseline. To better understand the reasons behind this performance gap between humans and machines, we took a closer look at the agents' answers. When considering the problem instances that were correctly solved by only one of the two agent types, the AI solved two to three times more instances than people. Moreover, when it comes to private data protection, the participants were much less likely to use data modifications that have high impact on prediction accuracy.
Overall, our results paint a rather dire picture when it comes to the potential use of human intuition to protect their private information from inference by AI.

Most of the works related to privacy protection focus on the role of a centralized authority in preserving the safety of information. Common notions of privacy include $k$\textit{-anonimity}~\cite{sweeney2002k}, which guarantees that any individual is indistinguishable from at least $k-1$ others, \textit{differential privacy}~\cite{dwork2008differential}, which guarantees that based on the output of an algorithm it is impossible to determine whether the data of a given individual was part of the input, and \textit{federated learning}~\cite{li2020federated}, where data is spread over multiple entities, and no one has access to the complete information. However, all these methods put the responsibility for protecting the private information of individuals in the hands of a central authority, which might be prone to error and negligence. In contrast, our work tests the ability of people to protect their own privacy.

The part of our study concerning gender and location prediction is closely related to the field of adversarial machine learning~\cite{huang2011adversarial}, which considers the process of doctoring inputs of machine learning algorithms. In particular, our study can be classified as part of the literature on evasion attacks~\cite{biggio2013evasion}, where adversarial modifications are introduced in testing instances, as opposed to poisoning attacks~\cite{chen2018automated}, where the training data gets altered. However, most of the literature considers the situation where the data is modified by algorithms, whereas we put this task in the hands of people.
A recent work evaluated the capability of social media users to understand what features of one's Twitter activity are most revealing to algorithms predicting our opinions~\cite{waniek2022hiding}. However, the authors did not test how modifying said features affects the accuracy of prediction, whereas we perform this evaluation in all settings that we consider.

The part of our study concerning the link prediction tasks is also related to the growing literature on strategically obscuring information from social network analysis tools. Some works proposed heuristic strategies of hiding certain relationships from link prediction algorithms based on the knowledge about local network neighborhood~\cite{zhou2019attacking,waniek2019hide}. A similar problem was considered for evading sign prediction algorithms, whose primary goal is to predict whether a particular link is positive or negative in a given social network~\cite{godziszewski2021attacking}. Other works have considered evading a variety of network analysis tools, including centrality measures~\cite{waniek2017construction,dey2019covert,waniek2020hiding,
was2020manipulability,waniek2021strategic,waniek2021members,waniek2022temporal}, community detection algorithms~\cite{waniek2018hiding}, and source detection algorithms~\cite{waniek2022social}. Compared to those works, the novelty of our approach lies in the examination of people's ability to infer private information, their ability to protect such information from inference attacks, and how their ability compares to that of algorithms designed specifically for this purpose.

We now discuss the policy implications of our results. The observed inadequacy of people to effectively protect private information from being inferred by AI underscores the need for new solutions to assist humans in these tasks. Otherwise, people may rely on their intuition, introducing certain modifications to their data before sharing it publicly, with the false belief that such modifications will safeguard them against attribute-inference attacks. As we have demonstrated, using algorithms is much more reliable than using one's intuition. This highlights the need to develop algorithms that can modify people's data, allowing them to share it with others, while ensuring that their secrets cannot be inferred from the shared content. Although people could outsource the task of data protection to a central authority, e.g., the company behind the social media platform they use, such an approach may be ineffective, as indicated by a plethora of privacy-related scandals. As such, users of the World Wide Web require tools and techniques that they can apply themselves to safeguard the information that they deem sensitive. Such tools might take the form of automated assistants, simple rule-of-thumb rules based on the inner working of the prediction algorithms, or entire applications devoted to the task of privacy preservation.

\section{Methods}
\label{sec:methods}

\subsection{Generating gender prediction instances}

To generate the gender instances, we use a dataset of about $81,000$ reviews left by the users of Yelp~\cite{reddy2016obfuscating}. Each review is labeled as posted by either a male or a female. We use the dataset to train an L2-regularized logistic regression classifier with bag-of-words count, following the work of Reddy and Knight~\cite{reddy2016obfuscating}. To generate the set of eye task instances, we select $1000$ reviews uniformly at random, and participants are asked to specify whether the review was written by a male or by a female.

We also generate $1000$ shield task instances based on reviews selected uniformly at random out of all reviews. We select four words of the review that are most indicative of the correct label, and four that are least indicative (according to the normalized pointwise mutual information~\cite{church1990word}). We allow the substitution of each of these eight words for its closest semantic equivalent computed using the \textit{word2vec} extension by Levy and Goldberg~\cite{levy2014dependency}. Participants are then asked to select three out of eight possible substitutions. To discern the participants' ability to solve the shield task, we ensure that the instances used in our experiments have a pool of answers that vary in terms of solution quality. More formally, in terms of the probability assigned to the correct gender by the prediction algorithm, we ensure that the possible solutions (i.e., the possible sets of three substitutions) include solutions that increase this probability by at least $0.025$, and also include solutions that decrease this probability by at least $0.025$. Moreover, we ensure that the standard deviation of the quality of possible solutions (i.e., the changes in the aforementioned probability) is at least $0.025$. The process of randomly selecting instances is repeated until all instances satisfy the above conditions.

\subsection{Generating location prediction instances}

To generate the location instances, we use a dataset of about $750,000$ photos downloaded from Flickr, provided by Yang et al.~\cite{yang2020protecting}. Each photo is labeled with the name of the country in which it was taken, selected from the following: Australia, Cambodia, Canada, China, Cuba, France, Germany, India, Ireland, Italy, Japan, Mexico, Netherlands, New Zealand, Norway, Peru, Portugal, Spain, Switzerland, Taiwan, Thailand, United Kingdom, and United States of America. Using this dataset, we train a deep convolutional neural network to produce, for any given photo, a probability distribution over all countries. To this end, we apply a standard VGG-16 architecture~\cite{vo2017revisiting}. Following Yang et al.~\cite{yang2020protecting}, the probability distribution for any given set of photos is generated as a product of the probability distributions corresponding to the photos in that set. Following Yang et al.~\cite{yang2020protecting}, each eye task instance consists of a set of $16$ photos taken in the same country, selected uniformly at random out of all photos taken in that country. When generating the instances, we ensure that they have the same distribution of countries as the original dataset; we ended up with $1010$ instances. Participants were then asked to select the country in which the photos were taken.

As for the shield task instances, we generate the same number of instances as the eye task instances (i.e., $1010$) with the same number of photos per instance (i.e., $16$), while ensuring that they have the same distribution of countries as the original dataset. Participants must select three photos to be removed from the set, with the goal being to hide the country in which the photos were taken, i.e., to minimize the probability assigned by the classifier to the correct country. To narrow down the possible photos that the participants can choose from, we allow them to choose from only eight photos, consisting of four photos whose removal results in the greatest drop in probability, as well as four photos whose removal results in the smallest drop. Again, to discern the participants' ability to solve the shield task, we ensure that the instances used in our experiments have a pool of answers that vary in terms of solution quality. That is, we ensure that the possible solutions include one that increases the probability by at least $0.025$, and another that decreases the probability by at least $0.025$, while also ensuring that the standard deviation of changes in probability is at least $0.025$.

\subsection{Generating link prediction instances}

To generate the networks that are part of the link prediction instances, we use three models, namely \BAn~\cite{barabasi1999emergence}, \ERn~\cite{erdds1959random}, and \WSn~\cite{watts1998collective}. We generate networks with $15$ nodes and an average degree of $4$. In the \WSn model, we set the rewiring probability parameter to $0.25$. We generated $334$ networks using each model, resulting in $1002$ networks.
For each network $G$ generated for the eye task, we select the node with the greatest degree as the evader $v^*$ (with ties being resolved uniformly at random) following Waniek et al.~\cite{waniek2019hide}. We then randomly select one of the links incident to the evader as the hidden link $e^*$, and remove it from the network. Participants are then presented with an image of network $G$ without the link $e^*$, and are asked to identify the hidden link.

For the shield tasks, we generate the same number of networks as for the eye task (i.e., $1002$), using the same three models, and select the link to be hidden (i.e., $e^*$) following the same steps mentioned above. Following Waniek et al.~\cite{waniek2019hide}, we consider the effectiveness of link prediction to be the best AUC (Area under the ROC curve) score of the following algorithms: common neighbours~\cite{newman2001clustering}, Salton~\cite{salton1986introduction}, Jaccard~\cite{jaccard1901etude}, S{\o}rensen~\cite{sorensen1948method}, hub promoted~\cite{ravasz2002hierarchical}, hub depressed~\cite{ravasz2002hierarchical}, Leicht-Holme-Newman~\cite{leicht2006vertex}, Adamic-Adar~\cite{adamic2003friends}, and resource allocation~\cite{zhou2009predicting}.
Out of all the links incident to either end of $e^*$, we select two links whose removal yields the greatest decrease in the effectiveness of link prediction, and two links whose removal yields the smallest decrease. Similarly, out of all the links that do not belong to $G$ and are incident to either end of $e^*$, we select two whose addition yields the greatest decrease in effectiveness, and two whose addition yields the smallest decrease. As a result, we end up selecting four links that can be removed from $G$, and four that can be added to $G$. Participants are then asked to select three out of these eight possible network modifications to execute in order to hide $e^*$ from AI. Again, to discern the participants' ability to solve the shield task, we ensure that the possible solutions include one that increases the AUC by at least $0.025$, and another that decreases the AUC by at least $0.025$, while also ensuring that the standard deviation of changes in AUC is at least $0.025$.

\section*{Ethics Statement}

The research was approved by the Institutional Review Board of the New York University Abu Dhabi (HRPP-2022-81). All research was performed in accordance with relevant guidelines and regulations. Informed consent was obtained from all participants.

\section*{Competing interest}

Authors declare no competing interest.

\section*{Data and code availability}

Data and code for reproducing the results of this study will be made available upon publication.

\bibliographystyle{abbrv}
\bibliography{bibliography-aishields}

\appendix
\clearpage

\section{Survey questionnaires}
\label{app:survey-questionnaires}

We first briefly describe the control flow of our experiment, before listing the entire text of the questionnaire.
Participating in the online survey consists of the following steps:

\begin{enumerate}

\item The participant is asked to accept the consent form (Section~\ref{app:survey-consent-form}).

\item The participant is asked to fill out a demographics questionnaire (Section~\ref{app:survey-demographics}).

\item The participant is randomly assigned to one of the six treatments:
\begin{itemize}
\item gender prediction eye task (Section~\ref{app:survey-gender-eye}),
\item location prediction eye task (Section~\ref{app:survey-location-eye}),
\item link prediction eye task (Section~\ref{app:survey-link-eye}),
\item gender prediction shield task (Section~\ref{app:survey-gender-shield}),
\item location prediction shield task (Section~\ref{app:survey-location-shield}),
\item link prediction shield task (Section~\ref{app:survey-link-shield}).
\end{itemize}
Below, the letter X stands for a section depending on the treatment under consideration.

\item The participant is presented with a set of instructions pertaining to the task (Section~A.X.1).

\item The participant is given a comprehension test about the task (Section~A.X.2).

\item If the participant manages to pass the comprehension test within two attempts, they are presented with five randomly selected instances of the task (Section~A.X.3); otherwise they are given a show-up fee.

\item After completing each instance, the participant is either:
\begin{itemize}
\item informed whether they correctly (Section~A.X.4) or incorrectly (Section~A.X.5) determined the required information for the eye tasks,
\item informed about their relative effectiveness (Section~A.X.4) for the shield tasks.
\end{itemize}

\item The participant is presented with a summary of payment (Section~\ref{app:survey-payment}).

\item The participant is asked to fill out a post-experimental questionnaire (Section~\ref{app:survey-post-experiment}).

\item The participant is thanked for participation and presented with a 10-digit validation code they have to enter in the MTurk system (Section~\ref{app:survey-end}).

\end{enumerate}

\subsection{Consent form}
\label{app:survey-consent-form}

Welcome to this study. which is carried out by Bedoor AlShebli, Marcin Waniek, and Talal Rahwan from New York university Abu Dhabi. 

Should you choose to participate in this study, you will do the following: 

\begin{enumerate}[noitemsep,topsep=0pt]
\item Fill out a brief questionnaire.
\item Read the instructions of the task you will be performing. 
\item Answer comprehension check questions to test your understanding of the task. If you fail twice to correctly answer all questions, you will not be allowed to finish the  study, but we will compensate you with \$0.10 tor your time. 
\item Complete the task.
\item Answer follow-up questions.
\end{enumerate}

You will receive a show-up fee of \$1.00 for completing the study, with the opportunity to  earn more, depending on your performance. The maximum bonus you may earn in the activity is \$2.5, In addition to your show-up fee.

The estimated time for the study is 12 minutes.

You are eligible to participate if you are:

\begin{enumerate}[noitemsep,topsep=0pt]
\item 18 years or older 
\item Live in the United States 
\end{enumerate}

Participation in this study is voluntary and you may leave the study at any point However, we can only pay you if you complete the study. When participating in this study we will keep track of your MTurk ID for payment and tracking. Information not containing identifiers may be used in future research or shared with other researchers without your additional consent. We do not anticipate any risks to you directly resulting from your participation in this study. There will also be no benefits to you beyond the money you earn completing the study. However, you will be assisting in the collection of valuable data about human behavior. 

It you have questions about either the study or your participation, you may contact Bedoor AlShebli at bedoor@nyu.edu. For questions about your rights as a research participant, you may contact the IRB and refer to \#HRPP-2019-93, New York University Abu Dhabi, +971 2628 4313 or IRBnyuad@nyu.edu. If you would like to have a copy of this document, please make a screenshot and keep it.

\textit{Important: by clicking the button below you agree to participate in the study. ONLY click this button it you intend to participate!}

\textit{You can only participate in the study once. Open ONLY ONE browser window and do not attempt to duplicate your input in any way otherwise you will not get paid.}

{[Button with an arrow pointing right]}

\subsection{Demographic questionnaire}
\label{app:survey-demographics}

\begin{itemize}

\item How old are you, in years?

[Text field accepting a number]

\item What is your gender? 

[Radio button with options ``Male'', ``Female'', and ``Other'']

\item With which of the following groups do you identify? You may select more than one.

[List of options: ``White'', ``Black/African-American'', ``Hispanic/Latino(a)'', ``Asian or Asian-American'', ``American Indian or Alaska Native '', ``Middle Eastern or North African'', and ``Other'']

\item In which state do you currently live? 

[Drop-down list with US states]

\end{itemize}

[Button with an arrow pointing left]

[Button with an arrow pointing right]

\subsection{Gender prediction eye task survey}
\label{app:survey-gender-eye}

\subsubsection{Gender prediction eye task instructions}

You will be presented with 5 reviews posted on the Internet. For each of them, you will be asked to determine whether the review was more likely to be posted by a female or a male*. For each review, you will receive a bonus of \$0.50 if you correctly determine the required information.

Here is an example of how the user interface will look like:

[Screenshot with an example of the interface]

The button allowing you to proceed to the next page will be activated after 30 seconds. Please, take your time to read the instructions carefully as a comprehension test will be shown on the next page.

\textit{* Note that the dataset we use comes from a study in which reviewers were classified based on their biological sex.}

[Button with an arrow pointing right]

\subsubsection{Gender prediction eye task comprehension test}

Please answer the following questions about the rules of the activity. If you fail twice to answer all of them correctly, you will not be able to participate in the study. You can review the instructions by clicking on the ``Instructions'' button at the bottom of the page. 

\begin{itemize}

\item What kind of information will be presented to you regarding each review? 
\begin{itemize}[noitemsep,topsep=0pt]
\item The text of the review, a photo of the product, and the name of the reviewer 
\item The text of the review and a photo of the product 
\item Only the text of the review 
\end{itemize}

\item What is the characteristic that you will be asked to determine for each review? 
\begin{itemize}[noitemsep,topsep=0pt]
\item The age of the reviewer 
\item The gender of the reviewer 
\item The education level of the reviewer 
\end{itemize}

\item How will your bonus be determined?
\begin{itemize}[noitemsep,topsep=0pt]
\item I will receive a bonus for any given review only if I correctly determine the reviewer's characteristic.
\item I will receive a bonus for every review, regardless of whether I correctly determine the reviewer's characteristic.
\item I will not receive any bonus.
\end{itemize}

\end{itemize}

[Button with ``Instructions'' label]

[Button with an arrow pointing right]

\subsubsection{Gender prediction eye task}

Task [the number of the task] of 5

Specify whether the person who posted the review below is more likely to be a female or a male.

[Text of the review]

[Radio button with options ``Male'', and ``Female'']

[Button with an arrow pointing right]

\subsubsection{Gender prediction eye task success}

Well done! You have correctly determined the gender Of the reviewer. As a result, you received a bonus of \$0.50

Your cumulative bonus thus far is: [Cumulative bonus value in USD]

[Button with an arrow pointing right]

\subsubsection{Gender prediction eye task failure}

Unfortunately, you have incorrectly determined the gender of the reviewer. The correct answer was [the correct answer]. As a result, you do not receive a bonus for this review.

Your cumulative bonus thus far is: [Cumulative bonus value in USD]

Here is a copy of the question: 

[The review text from the task]

[Button with an arrow pointing right]

\subsection{Location prediction eye task survey}
\label{app:survey-location-eye}

\subsubsection{Location prediction eye task instructions}

You will be presented with 5 sets, each consisting of sixteen photos that were taken in the same country. For each set of photos, you will be asked to specify the country in which the photos were taken, and you will receive a bonus of \$0.50 if you correctly determine the required information.

Here is an example of how the user interface will look like:

[Screenshot with an example of the interface]

The button allowing you to proceed to the next page will be activated after 30 seconds. Please, take your time to read the instructions carefully as a comprehension test will be shown on the next page.

[Button with an arrow pointing right]

\subsubsection{Location prediction eye task comprehension test}

Please answer the following questions about the rules of the activity. If you fail twice to answer all of them correctly, you will not be able to participate in the study. You can review the instructions by clicking on the ``Instructions'' button at the bottom of the page. 

\begin{itemize}

\item How many photos will each set consist of? 
\begin{itemize}[noitemsep,topsep=0pt]
\item 1 photo
\item 8 photos
\item 16 photos
\end{itemize}

\item What is the attribute that you will be asked to determine for each set of photos? 
\begin{itemize}[noitemsep,topsep=0pt]
\item The year in which the photos were taken 
\item The country in which all the photos were taken 
\item Whether the photos were taken by the same person 
\end{itemize}

\item How will your bonus be determined?
\begin{itemize}[noitemsep,topsep=0pt]
\item I will receive a bonus for any given set of photos only if I correctly determine the photos' attribute.
\item I will receive a bonus, regardless of my performance.
\item I will not receive any bonus.
\end{itemize}

\end{itemize}

[Button with ``Instructions'' label]

[Button with an arrow pointing right]

\subsubsection{Location prediction eye task}

Task [the number of the task] of 5

In which country do you think the following photos were taken? 

[A set of 16 photographs taken in the same country]

From [Drop-down list with countries]

[Button with an arrow pointing right]

\subsubsection{Location prediction eye task success}

Well done! You have correctly determined the country in which the photos were taken. As a result, you received a bonus of \$0.50

Your cumulative bonus thus far is: [Cumulative bonus value in USD]

[Button with an arrow pointing right]

\subsubsection{Location prediction eye task failure}

Unfortunately, you have incorrectly determined the country in which the photos were taken. The correct answer was [the correct answer]. As a result, you do not receive a bonus for this task. 

Your cumulative bonus thus far is: [Cumulative bonus value in USD]

Here is a copy of the question: 

[Picture of the set of photographs from the task]

[Button with an arrow pointing right]

\subsection{Link prediction eye task survey}
\label{app:survey-link-eye}

\subsubsection{Link prediction eye task instructions}

You will be presented with 5 social networks, each consisting of nodes (which represent people) and links (which represent friendships). In each network, two individuals know each other (meaning that there is supposed to be a friendship link between them), but they pretend that they do not know each other (meaning that the link between them is not part of the network). You will be asked to guess who these two individuals are, based on the structure of the social network. For each network, you will receive a bonus of \$0.50 if you correctly determine the required information. 

Here is an example of how the user interface will look like:

[Screenshot with an example of the interface]

The button allowing you to proceed to the next page will be activated after 30 seconds. Please, take your time to read the instructions carefully as a comprehension test will be shown on the next page.

[Button with an arrow pointing right]

\subsubsection{Link prediction eye task comprehension test}

Please answer the following questions about the rules of the activity. If you fail twice to answer all of them correctly, you will not be able to participate in the study. You can review the instructions by clicking on the ``Instructions'' button at the bottom of the page. 

\begin{itemize}

\item What does each network represent?
\begin{itemize}[noitemsep,topsep=0pt]
\item Streets within a city 
\item Connections between airports 
\item Friendships between people 
\end{itemize}

\item What is the information you will be required to determine for each network?
\begin{itemize}[noitemsep,topsep=0pt]
\item The number of links it has 
\item Which two nodes have an undisclosed link between them 
\item Which node is the most important 
\end{itemize}

\item How will your bonus be determined?
\begin{itemize}[noitemsep,topsep=0pt]
\item I will receive a bonus for any given network only if I correctly determine the required information.
\item I will receive a bonus for every network, regardless of whether I correctly determine the required information.
\item I will not receive any bonus.
\end{itemize}

\end{itemize}

[Button with ``Instructions'' label]

[Button with an arrow pointing right]

\subsubsection{Link prediction eye task}

Task [the number of the task] of 5

In this social network, two individuals know each other (meaning that there is supposed to be a friendship link between them), but they pretend that they do not know each other (meaning that the link between them is not part of the network). Based on the network structure, who are these two individuals? 

[Picture of a network]

From [Drop-down list with network nodes]

To [Drop-down list with network nodes]

[Button with an arrow pointing right]

\subsubsection{Link prediction eye task success}

Well done! You have correctly identified the two nodes that have an undisclosed link between them. As a result, you received a bonus of \$0.50

Your cumulative bonus thus far is: [Cumulative bonus value in USD]

[Button with an arrow pointing right]

\subsubsection{Link prediction eye task failure}

Unfortunately, you have incorrectly determined the two nodes that have an undisclosed link between them. The correct answer was [the correct answer]. As a result, you do not receive a bonus for this task. 

Your cumulative bonus thus far is: [Cumulative bonus value in USD]

Here is a copy of the question: 

[Picture of the network from the task]

[Button with an arrow pointing right]

\subsection{Gender prediction shield task survey}
\label{app:survey-gender-shield}

\subsubsection{Gender prediction shield task instructions}

You will be presented with 5 reviews posted on the internet. For each review, you will be informed about the gender of the person who wrote the review. Moreover, you will be asked to replace three words with their synonyms, in order to make it harder for an AI algorithm to correctly guess that person's gender*. To this end, you will be presented with eight word substitutions to choose from. For example, the substitution ``husband'' $\rightarrow$ ``hubby'' means that the word ``husband'' in the review will be replaced with the word ``hubby''. The harder you make it for an AI algorithm to correctly guess the reviewer's gender, the greater your bonus. For each review, the maximum bonus you may receive is \$0.50. 

Here is what the user interface will look like. As you can see, it displays a sample review, followed by eight suggested substitutions: 

[Screenshot with an example of the interface]

The button allowing you to proceed to the next page will be activated after 30 seconds. Please, take your time to read the instructions carefully as a comprehension test will be shown on the next page.

\textit{* Note that the dataset we use comes from a study in which reviewers were classified based on their biological sex.}

[Button with an arrow pointing right]

\subsubsection{Gender prediction shield task comprehension test}

Please answer the following questions about the rules of the activity. If you fail twice to answer all of them correctly, you will not be able to participate in the study. You can review the instructions by clicking on the ``Instructions'' button at the bottom of the page. 

\begin{itemize}

\item Your task is to make it harder for an AI algorithm to guess certain information. What is this information? 
\begin{itemize}[noitemsep,topsep=0pt]
\item The reviewer's age 
\item Whether review written by a human or by an AI
\item The reviewer's gender 
\end{itemize}

\item How will you perform this task? 
\begin{itemize}[noitemsep,topsep=0pt]
\item By adding two sentences to the review 
\item By replacing three words in the review 
\item By writing a completely new review 
\end{itemize}

\item How will your bonus be determined?
\begin{itemize}[noitemsep,topsep=0pt]
\item The harder I make it for an AI algorithm to correctly guess the information, the greater my bonus.
\item I will always receive a bonus, regardless of my performance. 
\item I will not receive any bonus.
\end{itemize}

\end{itemize}

[Button with ``Instructions'' label]

[Button with an arrow pointing right]

\subsubsection{Gender prediction shield task}

Task [the number of the task] of 5

[Text of the review]

The person Who wrote this review is [the gender of the author]. Which three substitutions do you propose in order to make it harder for an AI algorithm to correctly guess that person's gender? (Drag your choices into the box) 

Possible choices 
[List of eight possible modifications]

Your choices 
[List to which three modifications can be dragged]

[Button with an arrow pointing right]

\subsubsection{Gender prediction shield task feedback}

The effectiveness of your solution was [relative effectiveness of the participants answer] compared to the best solution. The best solution was [the best solution]. As a result, you received a bonus of [the bonus increment in USD]

Your cumulative bonus thus far is: [Cumulative bonus value in USD]

[Button with an arrow pointing right]

\subsection{Location prediction shield task survey}
\label{app:survey-location-shield}

\subsubsection{Location prediction shield task instructions}

A recent study has shown that when a person posts photos on the internet, an AI algorithm may be able to correctly guess the country in which these photos were taken.

You will be presented with 5 sets, each consisting of sixteen photos that were taken in the same country. For each set, you will be informed about the country in which the photos were taken. Moreover, you will be asked to remove three photos from the set, in order to make it harder for an AI algorithm to correctly guess the country. To this end, you will be presented with eight photos to choose from. The harder you make it for an AI algorithm to correctly guess the country in which the photos were taken, the greater your bonus. For each set of photos, the maximum bonus you may receive is \$0.50.

Here is what the user interface will look like. As you can see, it displays a sample set of sixteen photos; the photos you can choose from are highlighted by a red border, along with a letter associated with each photo: 

[Screenshot with an example of the interface]

The button allowing you to proceed to the next page will be activated after 30 seconds. Please, take your time to read the instructions carefully as a comprehension test will be shown on the next page.

[Button with an arrow pointing right]

\subsubsection{Location prediction shield task comprehension test}

Please answer the following questions about the rules of the activity. If you fail twice to answer all of them correctly, you will not be able to participate in the study. You can review the instructions by clicking on the ``Instructions'' button at the bottom of the page. 

\begin{itemize}

\item Your task is to make it harder for an AI algorithm to guess certain information. What is this information? 
\begin{itemize}[noitemsep,topsep=0pt]
\item The year in which the photos were taken 
\item The country in which the photos were taken 
\item Whether the photos were taken by a human or by an AI 
\end{itemize}

\item How will you perform this task? 
\begin{itemize}[noitemsep,topsep=0pt]
\item By adding three photos to the set 
\item By removing three photos from the set 
\item By adding a photo to and removing a photo from the set 
\end{itemize}

\item How will your bonus be determined?
\begin{itemize}[noitemsep,topsep=0pt]
\item The harder I make it for an AI algorithm to correctly guess the information, the greater my bonus.
\item I will always receive a bonus, regardless of my performance. 
\item I will not receive any bonus.
\end{itemize}

\end{itemize}

[Button with ``Instructions'' label]

[Button with an arrow pointing right]

\subsubsection{Location prediction shield task}

Task [the number of the task] of 5

The country in which these photos were taken is [the country name]. Out of the eight photos highlighted in red, which three do you propose we remove in order to make harder for an AI algorithm to correctly guess the country? (Drag your choices into the box) 

[A set of 16 photographs taken in the same country]

Possible choices 
[List of eight possible modifications]

Your choices 
[List to which three modifications can be dragged]

[Button with an arrow pointing right]

\subsubsection{Location prediction shield task feedback}

The effectiveness of your solution was [relative effectiveness of the participants answer] compared to the best solution. The best solution was [the best solution]. As a result, you received a bonus of [the bonus increment in USD]

Your cumulative bonus thus far is: [Cumulative bonus value in USD]

[Button with an arrow pointing right]

\subsection{Link prediction shield task survey}
\label{app:survey-link-shield}

\subsubsection{Link prediction shield task instructions}

A social network consists of nodes (which represent people) and links (which represent friendships). In such a network, it is possible that two individuals know each other (meaning that there is supposed to be a friendship link between them), but they pretend that they do not know each other (meaning that the link between them is not part of the network). A recent study has shown that, in such cases, an AI algorithm may be able to correctly guess who these two individuals are. 

You will be presented with 5 social networks. For each network, you will be informed about the identity of the individuals who pretend to not know each other. Moreover, you will be asked to introduce three modifications to the network, in order to make it harder for the Al algorithm to correctly guess who these two individuals are. To this end, you will be presented with eight possible modifications to choose from (these consist of four links that can be added to the network, and four links that can be removed). The harder you make it for an AI algorithm to correctly guess the individuals who pretend to not know each other, the greater your bonus. For each network, the maximum bonus you may receive is \$0.50.

Here is what the user interface will look like. As you can see, it displays a sample network. The yellow (dashed) link connects the two individuals who pretend to not know each other (you can see this link, but the Al algorithm cannot). The four blue (dotted) links can be added. The red (bold) links can be removed: 

[Screenshot with an example of the interface]

The button allowing you to proceed to the next page will be activated after 30 seconds. Please, take your time to read the instructions carefully as a comprehension test will be shown on the next page.

[Button with an arrow pointing right]

\subsubsection{Link prediction shield task comprehension test}

Please answer the following questions about the rules of the activity. If you fail twice to answer all of them correctly, you will not be able to participate in the study. You can review the instructions by clicking on the ``Instructions'' button at the bottom of the page. 

\begin{itemize}

\item Your task is to make it harder for an AI algorithm to guess certain information. What is this information? 
\begin{itemize}[noitemsep,topsep=0pt]
\item The number of links the network has 
\item Which node is the most important one in the network 
\item Which two individuals have an undisclosed link between them 
\end{itemize}

\item How will you perform this task? 
\begin{itemize}[noitemsep,topsep=0pt]
\item By adding five links
\item By adding or removing three links
\item By removing four link
\end{itemize}

\item How will your bonus be determined?
\begin{itemize}[noitemsep,topsep=0pt]
\item The harder I make it for an AI algorithm to correctly guess the information, the greater my bonus.
\item I will always receive a bonus, regardless of my performance. 
\item I will not receive any bonus.
\end{itemize}

\end{itemize}

[Button with ``Instructions'' label]

[Button with an arrow pointing right]

\subsubsection{Link prediction shield task}

Task [the number of the task] of 5

The yellow (dashed) link [the label of the hidden link] connects the two individuals who pretend to not know each other. The picture highlights eight possible modifications: four blue (dotted) links that can be added to the network, and four red (bold) links that can be removed from the network. Out of these eight modifications, which three do you propose in order to make it harder for an AI algorithm to correctly guess who these two individuals are?

[Picture of a network]

Possible choices 
[List of eight possible modifications]

Your choices 
[List to which three modifications can be dragged]

[Button with an arrow pointing right]

\subsubsection{Link prediction shield task feedback}

The effectiveness of your solution was [relative effectiveness of the participants answer] compared to the best solution. The best solution was [the best solution]. As a result, you received a bonus of [the bonus increment in USD]

Your cumulative bonus thus far is: [Cumulative bonus value in USD]

[Button with an arrow pointing right]

\subsection{Payment summary}
\label{app:survey-payment}

Thank you for your participation!

Your payment (including any bonuses you may have earned) is: [Total payment in USD]. 

You will receive your payment once you finish answering a few follow-up questions. Click "Continue" to answer a few questions about yourself and the task. 

[Button with ``Continue'' label]

\subsection{Post-experimental questionnaire}
\label{app:survey-post-experiment}

\begin{itemize}

\item What is your attitude towards Artificial Intelligence (Al)? 

[Radio button with options ``Extremely Positive'', ``Positive'', ``Slightly Positive'', ``Neutral'', ``Slightly Negative'', ``Negative'', and ``Extremely Negative'']

\item What is the highest Of education you have completed? 

[Radio button with options ``Less than high school'', ``High school or equivalent (e.g. GED)'', ``Some college'', ``2-year degree (Associate's)'', ``4-year degree (Bachelor's)'', ``Graduate or professional degree'']

\item What was your total family income from all sources in 2021, before taxes?

[Radio button with options ``Less than \$10,000'', ``\$10,000 to \$19,999'', ``\$20,000 to \$29,999'', ``\$30,000 to \$39,999'', ``\$40,000 to \$49,999'', ``\$50,000 to \$59,999'', ``\$60,000 to \$69,999'', ``\$70,000 to \$79,999'', ``\$80,000 to \$89,999'', ``\$90,000 to \$99,999'', ``\$100,000 to \$150,000'', ``More than \$150,000'']

\item Please, tell us your 5-digit zip-code. This information will only be used for statistical purposes.

[Text field accepting five digit code]

\item Was there anything noteworthy about the study? 

[Text field]

\item How would you rate your overall experience completing this study?

[Radio button with options ``Extremely Positive'', ``Positive'', ``Slightly Positive'', ``Neutral'', ``Slightly Negative'', ``Negative'', and ``Extremely Negative'']
\end{itemize}

[Button with an arrow pointing left]

[Button with ``Continue'' label]

\subsection{End screen}
\label{app:survey-end}

Thank you for your participation!

Your payment (including any bonuses you may have earned) is: [Total payment in USD]

Your validation code is: [10-digit validation code]

\clearpage
\section{Supplementary tables}
\label{app:supplementary-tables}

\begin{table}[tbh]
\centering
\begin{tabular}{llccc}
\toprule
Task & Attribute & Participants & \textit{t}-statistic & \textit{p}-value \\
\midrule
& Gender & $216$ & $9.719$ & $7.048 \times 10^{-22}$ \\
Eye & Location & $340$ & $10.074$ & $1.530 \times 10^{-23}$ \\
& Link & $200$ & $3.708$ & $2.152 \times 10^{-4}$ \\
\midrule
& Gender & $150$ & $21.779$ & $1.581 \times 10^{-91}$ \\
Shield & Location & $149$ & $54.786$ & $5.422 \times 10^{-313}$ \\
& Link & $110$ & $25.216$ & $8.290 \times 10^{-109}$ \\
\bottomrule
\end{tabular}
\caption{
\textbf{Statistical values corresponding to our experiment.}
The table specifies the number of participants who performed each task, as well as the results of the Welch's \textit{t}-test (the \textit{t}-statistic and the \textit{p}-value) comparing the performance of participants vs.~AI.
}
\label{tab:survey-stats}
\end{table}

\end{document}